\documentclass{iopjournal}
\usepackage{bm}
\usepackage{amsfonts}
\usepackage{amsmath}
\usepackage{multirow}
\usepackage{booktabs}
\usepackage{microtype}
\usepackage{ragged2e}
\usepackage{booktabs}
\usepackage{array}
\usepackage{caption}
\usepackage{array}
\begin{document}
\justifying
\articletype{Paper} %	 e.g. Paper, Letter, Topical Review...

\title{Network Knowledge Prior Guided Learning for Data-Efficient Surface Defect Detection}

\author{Hang-Cheng Dong$^{1,2}$\orcid{0000-0002-4880-6762},Guodong Liu$^{1,2}$,Dong Ye$^{1}$,Bingguo Liu$^{1,2,*}$}

\affil{$^1$School of Instrumentation Science and Engineering, Harbin Institute of Technology, Harbin, China}\\
\affil{$^2$Harbin Institute of Technology Suzhou Research Institute, Suzhou, China}

%\affil{$^*$Corresponding author.}

\begin{abstract}
Deep learning-based methods have become the de facto standard for industrial defect detection. However, their data-hungry nature and inherent ”black-box” characteristics often lead to performance bottlenecks and limited trustworthiness in real-world applications. To address these challenges, this paper proposes a novel knowledge-guided loss function that seamlessly integrates model interpretability into the training process without incurring any additional inference cost. Our method operates in two phases: first, a primary classification network is trained, and its explanations, in the form of saliency maps, are generated as prior knowledge. Second, a multi-task learning framework is established, where the main task performs classification, and an auxiliary task imposes consistency between the saliency maps of the final model and the primary model. This consistency is enforced by a dedicated knowledge-guided loss term, effectively acting as a powerful regularizer to steer the model towards robust feature representations. Extensive experiments on multiple public defect datasets demonstrate that our approach consistently enhances the performance of baseline models in terms of accuracy and AP. Moreover, visual analysis reveals that the proposed method yields more concentrated and human-intelligible saliency maps. This work presents a simple yet effective paradigm for bridging the gap between model performance and interpretability, paving the way for more reliable and high-performing vision systems in industrial quality inspection.
\end{abstract}

\vspace{6pt}

\section{Introduction}
\label{sec:introduction}
In modern manufacturing, measurement and inspection constitute a fundamental basis for product quality control and production process optimization. With manufacturing equipment evolving toward higher speed, higher precision, and greater intelligence, machine-vision-based automated visual inspection (AVI) has become an important alternative to manual inspection, enabling improved inspection efficiency and online quality monitoring capability \cite{fan2024novel,chen2024comparative,ping2023whole}. However, surface defects in real manufacturing scenarios are usually characterized by small scales, irregular morphologies, ambiguous boundaries, low contrast, and complex background textures. Variations in imaging conditions and process fluctuations further increase the difficulty of defect recognition. Therefore, industrial defect inspection requires not only determining whether defects are present, but also enabling models to extract stable and discriminative defect-related features from complex backgrounds.

Early industrial defect detection methods mainly relied on manual inspection and rule-based image processing algorithms. Manual inspection is inefficient and subjective, making it difficult to meet the requirements of stable inspection in high-speed continuous production. Rule-based methods generally depend on threshold segmentation, edge detection, texture analysis, and morphological operations. However, their performance is highly dependent on handcrafted feature design and parameter tuning. When product surface textures are complex, defect types are diverse, or imaging conditions vary significantly, fixed rules often fail to maintain satisfactory generalization performance \cite{chen2024comparative,lecun2015deep}.

In recent years, deep learning methods represented by convolutional neural networks (CNNs) have been widely applied to industrial visual inspection tasks \cite{lecun2015deep,weimer2016design}. These methods can automatically learn hierarchical feature representations from raw images in an end-to-end manner, reducing the reliance on handcrafted features and demonstrating strong discriminative capability in complex appearance defect recognition tasks. Existing deep-learning-based defect detection methods mainly include image-level classification, object detection, and semantic segmentation. Image-level classification methods can determine whether defects exist but cannot provide defect locations. Object detection methods can localize defect regions using bounding boxes, but they still face limitations when dealing with small-scale defects or defects with indistinct boundaries. Semantic segmentation methods can obtain more fine-grained defect regions, but they typically require a large number of pixel-level annotations, resulting in high annotation costs \cite{t6,t7}.

Although deep learning methods have significantly advanced defect detection technologies, most mainstream approaches still rely on fully supervised learning, which requires a large number of training samples with accurate category labels, bounding-box annotations, or pixel-level mask annotations \cite{t6,t7,t8}. However, in real industrial production, defect samples usually occur with low frequency, strong randomness, and long collection cycles, leading to limited data available for model training. Meanwhile, severe class imbalance often exists among different defect categories, and critical but low-frequency defect types are particularly difficult to collect sufficiently. For tiny defects, low-contrast defects, or defects with blurred boundaries, accurate annotation also requires domain experts, resulting in high cost, low efficiency, and subjective inconsistency. Therefore, sample scarcity, class imbalance, and high annotation costs constitute major limitations for the practical deployment of fully supervised defect detection methods in industrial scenarios.

To address these issues, existing studies have shown that multi-task learning and the incorporation of prior information are effective strategies for improving defect detection performance under limited annotation conditions. On the one hand, multi-task learning can jointly constrain related tasks such as classification, localization, segmentation, or reconstruction through a shared feature extractor. As a result, the model learns not only image-level discriminative results but also spatial response information related to defect regions. For example, a two-stage architecture combining a segmentation network and a decision network can enhance defect representation capability when only a small number of defect samples are available \cite{t6}. Similarly, multi-task frameworks involving defect restoration and ROI generation have demonstrated that auxiliary tasks help suppress noise from non-defective regions and enhance responses in defect-related regions \cite{t7}. On the other hand, weakly supervised labels, class-aware heatmaps, regions of interest (ROIs), and knowledge from teacher models can provide additional spatial constraints for model training without substantially increasing manual annotation costs, thereby improving the model’s ability to learn defect locations and discriminative regions \cite{t7,t8}. Therefore, combining multi-task constraints with available prior information is beneficial for improving model generalization and detection robustness under limited-sample conditions.

In addition, under limited-sample and high-noise conditions, deep learning models are prone to overfitting \cite{t9}. A model may not truly learn defect-relevant discriminative features, but instead rely on background textures, illumination variations, or incidental correlations introduced during data acquisition for classification. Although such spurious correlations may lead to high accuracy on the training set, they can easily cause performance degradation when the model is applied to new production batches, imaging devices, or process environments \cite{t6,t9}. Meanwhile, the decision-making process of deep learning models is usually difficult to interpret explicitly, and their outputs cannot be directly associated with specific defect regions or process factors. This “black-box” nature reduces quality engineers’ trust in inspection results and increases the difficulty of error tracing and model debugging \cite{t10,t11}.

To alleviate the opacity of model decision-making, explainable artificial intelligence (XAI) methods have gradually been introduced into industrial defect detection tasks \cite{t10,t11}. Among them, saliency maps or attention maps are commonly used visual explanation techniques. For example, the Grad-CAM algorithm generates class activation heatmaps based on the gradient responses of target categories, thereby visualizing the image regions that the model focuses on during classification decision-making \cite{t12}. Through saliency maps, it is possible to determine whether the model attends to true defect regions or is instead disturbed by non-defective factors such as background textures, edges, and noise.

However, most existing saliency-map-based methods are used as post-hoc explanation tools after model training. Their primary role is to explain decisions that have already been formed by the model, rather than directly participating in the training process. Therefore, the defect-region responses and spatial attention distributions contained in saliency maps have not yet been fully transformed into effective supervisory signals for training. For industrial defect detection tasks with scarce defect samples and high costs of precise annotation, if model-generated saliency maps can be introduced into the training process as weakly supervised spatial priors, it may be possible to guide the model toward more discriminative defect regions without increasing manual annotation costs.

Knowledge distillation (KD) provides another perspective for model knowledge transfer and training regularization. Conventional knowledge distillation usually transfers class probabilities, intermediate features, or spatial response information from a teacher model to a student model, thereby improving the discriminative capability and generalization performance of the student model \cite{t13,t14}. However, in industrial defect detection tasks, how to transform explainability results such as saliency maps into spatial constraints during training, while simultaneously improving model performance and explanation reliability, remains an open problem worthy of further investigation \cite{t8,t13,t14}.

Based on the above analysis, this paper proposes a knowledge-guided learning framework that transforms saliency maps from conventional post-hoc explanation results into spatial prior information during training. The proposed method consists of two stages. The first stage is knowledge generation. Specifically, a standard classification network is first trained on the defect dataset, and sample-level saliency maps are then generated using the trained model and stored as prior knowledge. The second stage is knowledge-guided learning. In this stage, a model with the same network architecture is retrained, where the primary task is defect classification, and an auxiliary constraint is introduced to measure the discrepancy between the saliency map of the current model and the prior saliency map. By constructing a knowledge-guided loss function, the model optimizes classification performance while its spatial response distribution is constrained by saliency-map priors. This reduces the model’s dependence on background textures, noise regions, and non-discriminative features, guiding it to focus more stably on defect-related regions.

In summary, the main contributions of this paper are as follows:

{\small $\bullet$}{ A knowledge-guided loss function is proposed, which introduces model-generated saliency maps as spatial priors without requiring additional annotations. The proposed loss regularizes the feature attention regions of the defect detection model during training.}
% labor-intensive problem 

{\small $\bullet$} We propose a two-stage multi-task learning framework for knowledge-guided training based on saliency-map consistency constraints. The proposed framework is independent of specific network architectures and introduces no additional parameters or computational overhead during inference.

{\small $\bullet$} We propose an explainability knowledge injection mechanism for defect detection tasks. By transforming saliency maps from conventional post-hoc diagnostic tools into active guidance information during training, the proposed mechanism enables explainability information to participate in the feature learning process while the model learns discriminative features for defect classification.

\section{Related Work}

This study lies at the intersection of deep-learning-based defect detection, model explainability, and knowledge-aware model training. Therefore, this section reviews related studies from these three perspectives and further clarifies the connections and differences between existing methods and the method proposed in this paper.

\subsection{Deep-Learning-Based Defect Detection}

Deep learning has become an important methodology in the field of surface defect detection. According to different task formulations, existing methods can be broadly categorized into defect classification, object detection, and defect segmentation. Defect classification methods mainly determine whether defects are present in an image or identify the corresponding defect categories, making them suitable for rapid quality assessment. For example, Liu et al. proposed a deep learning model for imbalanced multi-label data to address the class imbalance problem in steel surface defect classification \cite{l1}. Konovalenko et al. adopted a deep residual network for steel surface defect classification and demonstrated the effectiveness of deep feature representations in recognizing complex defects \cite{l2}. In addition, He et al. further combined semi-supervised learning with generative adversarial networks to alleviate the shortage of labeled samples in steel surface defect classification \cite{l3}. However, classification methods usually provide only image-level predictions and are unable to indicate the specific locations and spatial extents of defects.

To meet the requirements of defect localization and regional analysis in industrial applications, object detection and segmentation methods have also been widely investigated. Object detection methods generally localize defects using bounding boxes and provide a favorable trade-off between detection accuracy and inference speed. For example, Chen et al. proposed ESDDNet, which enhances the detection capability for small defects on workpiece surfaces through parallel convolution, serial convolution, and feature fusion modules \cite{l4}. Fan and Qiu developed a deep-learning-based detection algorithm for surface defects in injection-molded products to improve the accuracy and efficiency of machine vision inspection \cite{l5}. Zhang et al. proposed CADN, which achieves defect classification and localization using only image-level labels, thereby reducing the requirement for precise annotations \cite{l6}. In contrast, segmentation methods can obtain more fine-grained pixel-level defect regions and are suitable for evaluating defect area, morphology, and severity. Tabernik et al. proposed a surface defect detection method based on a segmentation network and a decision network, achieving satisfactory detection performance under limited defect samples \cite{l7}. Neven and Goedemé proposed a multi-branch U-Net architecture to simultaneously segment defect types and severity levels on steel surfaces \cite{l8}.

Overall, existing methods have gradually evolved from image-level classification to object detection and pixel-level segmentation. Detection performance has been improved through strategies such as multi-scale feature fusion, weakly supervised learning, lightweight network design, and segmentation-based architectures \cite{l4,l5,l6,l7,l8}. However, most of these methods remain highly dependent on data-driven learning and still face several challenges in real industrial scenarios, including defect sample scarcity, imbalanced class distributions, complex background interference, and high annotation costs. Different from approaches that improve performance by modifying network architectures or introducing additional data augmentation strategies, this paper focuses on training constraints. Without changing the backbone network structure or the input data format, a knowledge-guided loss function is introduced, enabling standard CNN models to receive additional spatial prior constraints beyond classification supervision. This improves the robustness and defect-discriminative capability of the model.

\subsection{Model Explainability in Visual Inspection}

The black-box nature of deep models has promoted the development of explainable artificial intelligence (Explainable AI, XAI) in visual inspection tasks. Among existing XAI techniques, saliency visualization methods based on class activation mapping (CAM) are among the most commonly used. CAM was first proposed to generate class-discriminative heatmaps using global average pooling and classification weights, thereby endowing classification networks with a certain degree of weak localization capability \cite{z1}. Subsequently, Grad-CAM generated heatmaps by exploiting the gradients of a target class with respect to convolutional feature maps, making this approach applicable to a broader range of CNN architectures \cite{t12}. Based on this idea, Grad-CAM++ improved the calculation of gradient weights and enhanced localization performance in multi-object or small-object scenarios \cite{z3}. Score-CAM removed the dependence on gradient information by weighting activation maps according to forward-propagation scores, thereby improving the stability and interpretability of the generated heatmaps \cite{z4}. LayerCAM further utilized hierarchical response information from different convolutional layers to enhance the representation of fine-grained spatial regions in heatmaps \cite{z5}. In addition, Ablation-CAM evaluated the contribution of feature maps to classification results through channel-wise ablation, providing another gradient-free perspective for explainability analysis \cite{z6}.

In defect detection tasks, CAM-based methods have been used not only for post-hoc model interpretation but also increasingly for weakly supervised localization and pseudo-label generation. Previous studies have introduced Grad-CAM into surface defect detection models to localize defect-related regions through heatmaps, thereby assisting in verifying the rationality of model classification results \cite{z7}. Furthermore, RA-CAM improves the resolution of CAM heatmaps through filtered guided backpropagation and region-aware weighting, and uses the generated heatmaps as pseudo-labels for weakly supervised defect segmentation \cite{z8}. LTGrad-CAM enhances the fine-grained localization capability of CAM-based methods in weakly supervised defect detection by integrating shallow high-resolution semantic information and suppressing background noise through gradient truncation \cite{z9}. These studies indicate that CAM-based heatmaps have gradually evolved from simple visual explanation tools into important sources of spatial information for weakly supervised defect localization and segmentation.

However, existing CAM-based methods are still mainly used for post-hoc analysis, weakly supervised pseudo-label generation, or visual verification of defect regions. Their explanation results are usually generated after model training and do not directly participate in the optimization process of the original classification model. In other words, although CAM-based heatmaps can reveal the regions attended to by the model, the spatial response information contained in these heatmaps has not been further transformed into training supervision signals for the classification network itself. In contrast, this paper transforms saliency heatmaps from passive diagnostic tools into active guidance signals during training, thereby establishing a connection between model interpretability and model performance improvement.

\subsection{Multi-Task Learning and Regularization}

Multi-task learning (MTL) is a classical learning paradigm whose core idea is to share feature representations across multiple related tasks and improve model generalization by exploiting the inductive bias provided by auxiliary tasks \cite{x24}. In computer vision, common multi-task learning frameworks usually center on a primary task and jointly optimize auxiliary tasks such as semantic segmentation, depth estimation, or instance segmentation, enabling the model to learn more robust feature representations under the joint constraints of different supervision signals \cite{x25}.

The framework proposed in this paper can also be regarded as a new form of multi-task learning. Specifically, the primary task is defect classification, while the auxiliary task is saliency heatmap regression or saliency consistency constraint. It is worth noting that this auxiliary task does not require additional manual annotations, since its supervision targets are automatically generated by a pretrained model. Similar to the idea of transferring output distributions, intermediate features, or attention information from a teacher model to a student model in knowledge distillation \cite{x26,x27}, this paper uses saliency heatmaps generated by a pretrained model as prior knowledge to provide additional spatial constraints for subsequent model training.

From the perspective of regularization, the proposed knowledge-guided loss can be viewed as a task-related spatial regularization term. By penalizing model attention that is inconsistent with the prior saliency information, it constrains the model hypothesis space, thereby reducing the risk of overfitting and guiding the network to learn more discriminative and semantically reasonable features. This method is clearly different from conventional regularization techniques, such as norm regularization or dropout. Traditional regularization methods mainly suppress overfitting through parameter constraints, noise injection, or stochastic deactivation \cite{x28,x29}, whereas the proposed method introduces high-level spatial constraints that are generated from the data itself and are related to defect regions.

\section{Methodology}
Deep learning models, especially deep convolutional neural networks (CNNs), have achieved remarkable breakthroughs in visual tasks such as image classification, object detection, and semantic segmentation. This success is largely attributed to their end-to-end feature learning capability, which enables them to automatically extract complex hierarchical representations from large-scale data. However, the black-box nature of deep models makes their decision-making mechanisms difficult to interpret, raising concerns regarding model reliability, fairness, and safety.

At present, methods for improving the classification performance of deep neural networks mainly focus on network architecture design, optimization strategy improvement, and data augmentation. Although these methods can enhance model representation capability and generalization performance, they usually do not explicitly exploit or constrain the internal knowledge representations formed during model training. Saliency-map-based visual explanation techniques provide an important means for observing and verifying the internal knowledge of models. However, existing studies mostly treat explainability as a post-training tool for analysis, diagnosis, or debugging, while rarely integrating it into the training process to actively guide the formation of knowledge representations. Therefore, this chapter proposes a knowledge-guided loss function framework, aiming to promote the joint improvement of classification performance optimization and explainability analysis.

\subsection{Characteristics of Defect Detection Tasks}

The particular challenges of defect detection tasks mainly arise from multiple constraints related to data distribution, defect morphology, and industrial application scenarios. These characteristics can be summarized into four aspects: extreme data imbalance and defect rarity, high variability and subtlety of defect appearances, complex background interference and diversity of normal samples, and strict requirements for decision reliability and interpretability.

First, in normal production processes, qualified products account for the vast majority, while defective samples occur at a very low frequency. This leads to severe class imbalance in training datasets. As a result, models can be easily dominated by massive normal samples and may learn a trivial strategy of simply predicting all samples as normal, thereby failing on rare defective samples. More importantly, the limited number of defect samples is often insufficient to cover all possible anomaly patterns, resulting in poor generalization to unseen defect types. This characteristic requires models to have a strong ability to learn discriminative features from few and uncommon samples while effectively suppressing bias toward the majority class.

Second, industrial defects exhibit extremely diverse morphologies, scales, and visual appearances. Even defects belonging to the same category, such as scratches, may vary significantly in length, width, orientation, contrast, and boundary shape. Meanwhile, many critical defects, such as tiny surface flaws or early-stage cracks, show only subtle visual differences from normal backgrounds and often have a low signal-to-noise ratio. This is in sharp contrast to natural images, where objects usually possess relatively consistent semantic structures and salient visual appearances. Therefore, defect detection models cannot rely solely on fixed and coarse-grained feature patterns. Instead, they must be capable of sensitively capturing tiny, local, and highly variable abnormal signals, which imposes high requirements on the granularity of feature extraction.

In addition, normal surfaces of industrial products may themselves contain complex textures, structural patterns, illumination variations, or legitimate surface patterns. Such background information acts as strong noise and can easily be confused with real defects. For example, a bright texture edge may be mistakenly identified as a scratch. Therefore, models must accurately distinguish inherent variations in background textures from true anomalies. This requires the model to understand the distribution boundary of normal samples and learn representations that are highly robust to background interference.

Finally, industrial defect detection involves high decision-making risk. Missed detections may allow defective products to enter subsequent production stages or even reach the market, causing safety hazards and reputational losses. False alarms may lead to unnecessary product scrapping, rework, or production interruption, increasing economic costs. Therefore, models should not merely output simple classification results or confidence scores, but should also provide understandable evidence for their decisions. An interpretable model can indicate the regions it regards as defective, providing direct support for manual verification and process improvement. This is a critical requirement for industrial deployment.

\subsection{Deep Learning Explainability and Network Knowledge Extraction}

The interpretability of deep learning models aims to reveal the logic and evidence underlying their internal decision-making processes. In image classification tasks, visual explanation is one of the most intuitive forms of interpretability. Its objective is to generate heatmaps, also known as saliency maps, that highlight the regions in the input image that play a critical role in the model’s prediction. To extract reliable and structured “model knowledge” from a trained baseline classification model, we adopt and systematically investigate two advanced visual explanation methods: FullGrad \cite{srinivas2019full} and LayerCAM \cite{jiang2021layercam}. These two methods represent two important directions, namely gradient-complete attribution and flexible class activation map generation. Their principles and characteristics jointly constitute the technical foundation of the proposed knowledge extraction stage.

The core idea of FullGrad originates from a critical reconsideration of the limitations of early gradient-based attribution methods. Traditional methods such as Vanilla Gradient mainly attribute model predictions to variations in input pixels, while ignoring a fundamental component of deep neural network architectures: the bias terms. FullGrad points out that, for a deep network using ReLU-like activation functions, the network output can be exactly decomposed into the sum of contributions from input pixels and contributions from all bias terms inside the network. Therefore, a truly complete attribution explanation must include both parts; otherwise, the explanation is inherently incomplete.

From a mathematical perspective, FullGrad is built upon a rigorous gradient decomposition. For a deep network $f$, its output $f(x)$ can be expressed as:

\begin{equation}
f(x) = \sum_i \frac{\partial f}{\partial x_i} gx_i + \sum_k \frac{\partial f}{\partial b^k} gb^k
\label{eq:fullgrad_decomposition}
\end{equation}

Here, the first term represents the contribution of input pixels, while the second term represents the contribution of all internal bias terms b
k
. The generation of FullGrad saliency maps can be regarded as a visual implementation of this decomposition. Specifically, FullGrad first computes the gradients of the output with respect to the input pixels and intermediate-layer activations. It then performs element-wise multiplication between the input gradients and the input values, while also multiplying the activation gradients of each layer with the corresponding activation values. The latter essentially encodes the contribution of the bias terms in that layer. Finally, the attribution maps from all layers are upsampled to the input spatial resolution and added to the input attribution map, followed by necessary post-processing, thereby generating the final attribution heatmap that satisfies the completeness axiom.

In general, the characteristics of FullGrad make it an ideal tool for knowledge extraction. First, its theoretical completeness ensures the reliability and systematic nature of the attribution results, and the residual-free decomposition provides a solid mathematical foundation for explanation. Second, its structure-aware capability can reveal the influence of internal network parameters, such as bias terms, on model decisions, thereby offering deeper structural insights. Finally, by integrating information from different layers of the network, the saliency maps generated by FullGrad are usually more stable, smoother, and more comprehensive. These characteristics indicate that the “knowledge” extracted from the baseline model is global and trustworthy. When used as a constraint, such knowledge can guide a new model to learn a structurally more reasonable decision-making pattern with a more balanced contribution distribution.

LayerCAM was proposed to address the limitations of the classical Grad-CAM family when exploiting shallow network features. Grad-CAM obtains channel weights through global average pooling over gradients. Although this operation works well for deep semantic features, it severely loses the rich spatial details contained in shallow feature maps, resulting in blurry saliency maps and coarse localization. The core innovation of LayerCAM is that it abandons global pooling and instead adaptively preserves and utilizes gradient information in the spatial dimension. As a result, it can generate high-quality and high-resolution class activation maps from any layer of the network.

Its working principle is as follows. For a target class \(c\) and a selected layer \(l\) of the network, LayerCAM first computes the gradient \(G^l\) of the class score with respect to the feature map \(A^l\) of that layer. The key step is that it does not perform global pooling on the gradients. Instead, it applies the ReLU function to the gradient values at each spatial location and channel, obtaining spatially adaptive weights:

\begin{equation}
M_{\mathrm{Layer\text{-}CAM}}^{c}
=
\mathrm{ReLU}
\left(
\sum_{k} w_{l}^{ck} \otimes A_{l}^{k}
\right)
\end{equation}

This weight map explicitly reflects the importance of each spatial location for the class prediction. Subsequently, these spatial weights are used to weight the original feature maps channel by channel. All weighted feature maps are then summed and passed through a ReLU activation to obtain the class activation map of the selected layer. To obtain a more comprehensive saliency map that integrates multi-scale information, activation maps from shallow, intermediate, and deep layers can be upsampled respectively and then aggregated through element-wise maximum operation or other fusion strategies.
\begin{figure}[]
	\centering
	\includegraphics[width=\linewidth]{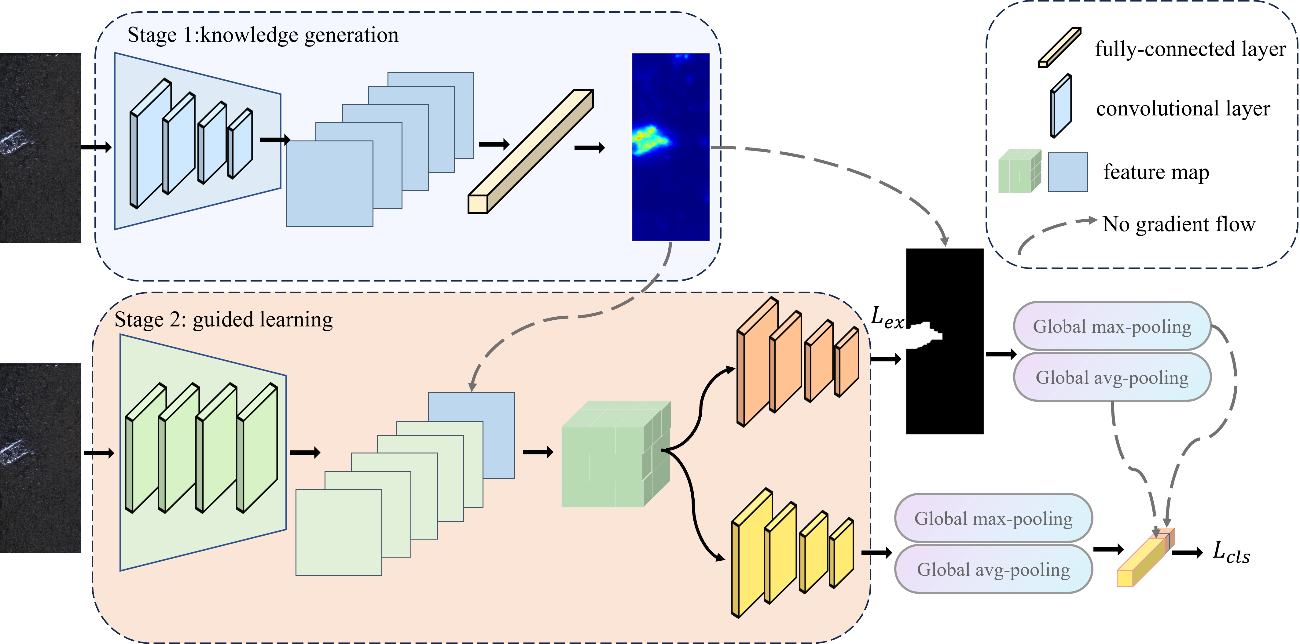}
	\caption{Linear approximation, polynomial approximation, and piecewise linear approximation}
	\label{Figure.1.1}
\end{figure}

Similar to FullGrad, LayerCAM also exploits layer-wise feature information, but the extracted knowledge differs in nature. Owing to its high-resolution representation and accurate localization capability, the knowledge extracted by LayerCAM can clearly delineate the details and boundaries of target regions, which is critical for improving the discriminative power of the classification model. Its layer flexibility allows us to freely explore and fuse attention patterns at different abstraction levels of the network. Ultimately, the saliency map generated through multi-scale fusion contains hierarchical knowledge ranging from local textures to global semantics. Such fine-grained spatial guidance can effectively encourage the new model to establish a coherent attention mechanism from local details to the overall structure during training, thereby learning more robust feature representations.

\subsection{Explainability-Guided Feature Learning Network}

CNN is a fundamental deep learning architecture mainly built upon convolution operations, and it is widely used in image- and video-based detection tasks. Defect detection requires the analysis of acquired defect sample images. However, traditional image processing methods often struggle with variations in environment, illumination, and background conditions. Therefore, CNNs are adopted in this study, as they are more suitable for handling such complex visual inspection problems. The overall design framework proposed in this section is shown in the figure \ref{Figure.1.1}.

\subsection{Backbone Convolutional Neural Network Design}

Convolution operations can be used to construct deep convolutional neural networks, enabling effective representation of image features. In traditional image processing, convolution kernel weights are usually manually designed to extract specific local features, which belongs to handcrafted feature engineering. In contrast, convolutional neural networks do not predefine fixed convolution kernel parameters. Instead, these parameters are automatically updated during training through backpropagation, thereby enabling data-driven feature extraction. In general, the strong feature representation capability of CNNs is mainly attributed to this automatic learning mechanism.

In addition, convolutional neural networks usually build deep architectures by stacking multiple convolutional modules. Deeper networks can extract more complex and abstract feature representations. However, when the network becomes excessively deep, optimization difficulties may lead to degradation, where model performance no longer continues to improve as the number of layers increases. Based on this consideration, this study draws on the network design ideas of VGG and ResNet, and adopts a multi-layer convolutional structure as the feature extraction backbone. 

\subsection{Network Knowledge Extraction and Prior Knowledge Injection}

Classical deep learning methods usually adopt an end-to-end training strategy. For defect recognition tasks, when sufficient training data and rich annotation information are available, convolutional neural networks can generally achieve satisfactory recognition performance. However, in practical industrial scenarios, defect samples are often scarce and annotation costs are high. Both the amount of data and the amount of annotation may therefore be limited, which restricts the full exploitation of the feature learning capability of convolutional neural networks.

To address the above issues, this study proposes a guided learning method based on knowledge priors. The main characteristic of this method is that it does not require additional manual annotations. Instead, the explainability results generated by the model itself are used as knowledge priors to inversely guide subsequent model training. Meanwhile, the proposed method mainly acts during the training stage, introduces almost no additional inference cost, and only brings a small amount of extra computational overhead.

The proposed knowledge-guided learning method adopts a two-stage training strategy. In the first stage, a conventional convolutional neural network, such as VGG16 or ResNet, is trained. After training, explainability methods such as FullGrad, LayerCAM, or Grad-CAM are used to generate saliency heatmaps. In this study, the saliency heatmap is considered to reflect, to a certain extent, the key regions on which the current model decision depends. Therefore, it is defined as the network knowledge prior.

In the second stage, the above network knowledge prior is introduced into the model training process to constrain and guide the model to focus on more discriminative image regions. Specifically, for each training sample x, an attribution saliency map H(x) is generated using LayerCAM and FullGrad. On the one hand, as shown in the figure, the attribution saliency map H(x) is processed through image processing operations and then used as a prior shape constraint. As a part of the loss function, this prior constraint is used to regularize model learning, which will be introduced in the next subsection.

On the other hand, the attribution saliency map $H(x)$ is directly treated as a channel of information and concatenated with the output $F(x)$ of the backbone network, yielding a feature map 
$P(x) = [F(x), H(x)]$ that incorporates prior knowledge. 
$P(x)$ will provide a basis for subsequent network decision. 
Furthermore, $P(x)$ is fed into a segmentation subnetwork, and the output is denoted as $\mathrm{Seg}(P(x))$. Similarly, the features containing the defective target regions are injected via pooling into the fully connected layer of the classification head to guide and constrain the classifier's performance.

This study adopts two pooling methods to extract features: Global Max Pooling and Global Average Pooling, denoted as GMP and GAP respectively. Global max pooling highlights the most discriminative local features but completely discards global spatial distribution information. Global average pooling computes the average value of the feature map, thus preserving global context information. Using both pooling methods simultaneously can efficiently convey the defect representations from the attribution saliency map to the classifier. Finally, we concatenate $\mathrm{GMP}(\mathrm{Seg}(P(x)))$ and $\mathrm{GAP}(\mathrm{Seg}(P(x)))$ with the MLP features before the final classification decision, termed knowledge injection:

\begin{equation}
y = \text{MLP}([\text{GMP}(\text{Seg}(P(x))), \text{GAP}(\text{Seg}(P(x))), \text{MLP}_{\text{feat}}])
\end{equation}

where $y$ denotes the classifier's logit output, i.e., the unnormalized network output.

In summary, this study designs two feature injection mechanisms, applied to the convolutional layers and the fully connected layers respectively, and preserves image features to the fully connected layer via global max pooling and global average pooling. Thanks to the prior knowledge of the network, additional semantic information is gained with only a small training overhead. This efficient feature extraction design is particularly suitable for the characteristics of defect detection tasks, enhancing detection performance under conditions of data scarcity, large defect variations, and complex backgrounds.

\subsection{Knowledge-Guided Multi-Task Loss Function Design}

To effectively train the proposed guided learning framework, this study adopts a multi-task loss function design. In previous research practices, multi-task learning has been recognized as a means to improve the expressiveness of shared features, thereby enhancing learning efficiency and accuracy. Following this design principle, we formulate two task losses: (1) a primary classification loss, and (2) a guiding task, which employs a segmentation loss without requiring segmentation labels. Moreover, the segmentation subnetwork and the classification subnetwork are trained relatively independently, meaning that the gradient flow from the classification subnetwork does not propagate into the segmentation subnetwork.

For both the classification loss and the segmentation loss, this study adopts the cross-entropy loss function to optimize the network parameters, as the segmentation task is essentially a pixel-wise classification problem. The cross-entropy loss function originates from information theory and measures the discrepancy between two probability distributions. Specifically, when the number of classes \( n = 2 \), the Binary Cross Entropy (BCE) loss is defined as:

\begin{equation}
\mathcal{L}_{\text{BCE}} = -\frac{1}{N} \sum_{i=1}^{N} \left[ y_i \log(\hat{y}_i) + (1 - y_i) \log(1 - \hat{y}_i) \right]
\end{equation}

where $m$ denotes the number of samples, $y_i$ is the ground-truth label of the sample, and $\hat{y}_i$ is the model prediction. The segmentation loss function $\mathcal{L}_{\text{seg}}$ adopts the same form, where $y_i$ represents the ground-truth category label of each pixel. The segmentation loss is calculated by traversing all pixels in the target class regions and then taking the average value. However, pixel-level ground-truth annotations are not provided in this study. Therefore, the attribution saliency heatmap is used to generate pseudo-labels as approximate ground truth.

Specifically, the detailed procedure of the segmentation loss function is  $\mathcal{L}_{\text{seg}}$ is illustrated in the figure. For the attribution saliency map $H(x)$ generated by the explainable method, a pseudo-label $H'(x)$ is obtained through image binarization. The binarization threshold is determined using the classical adaptive thresholding method, namely Otsu's method.

The total loss integrates both the classification loss and the segmentation loss. However, considering that the segmentation loss is not completely accurate and can only serve as auxiliary information, a coefficient is assigned to the segmentation loss so that its weight gradually decreases as training proceeds. Therefore, the total loss function can be written as:

\begin{equation}
L_{\text{total}} = \lambda L_{\text{seg}} + (1 - \lambda) L_{\text{cls}}
\end{equation}

where $\lambda$ denotes the weight coefficient, which is computed as:

\begin{equation}
\lambda = 1 - \frac{k}{k_{\text{epoch}}}
\end{equation}

where $k$ is the current training epoch and $k_{\text{epoch}}$ is the total number of training epochs.

In summary, this study adopts two levels of supervisory signals simultaneously. The segmentation task, serving as a guided learning task, provides pixel-wise supervision to assist the classification task in learning more discriminative representations. As training proceeds, the classification loss gradually becomes the dominant component of the total loss, thereby avoiding overfitting to erroneous segmentation features.

\section{Experiments}

\subsection{Datasets Descriptions}

We evaluated the proposed weakly supervised semantic segmentation method on two widely used defect detection datasets. The specifications of these datasets are detailed as follows.

(1) KSDD dataset.

The KolektorSDD dataset \cite{q1}, developed through collaboration with the Kolektor Group, focuses on electrical commutator surface defects with expert annotations. Original image dimensions maintain a fixed width of 500 pixels while exhibiting height variations between 1240 and 1270 pixels. This benchmark contains 399 annotated samples, comprising 52 defective instances and 347 defect-free cases, systematically collected from 50 distinct physical commutator units.

(2) KSDD2 dataset.

The second dataset, KSDD2 (Kolektor Surface Defect Detection Dataset Version 2) \cite{q2}, comprises 3,012 high-resolution industrial inspection images with nominal dimensions of 230×630 pixels. This industrial-grade dataset is partitioned into a training set containing 2,085 defect-free samples (246 defective cases) and a test set comprising 894 non-defective specimens (110 defective instances), maintaining an 8:2 split ratio between training and evaluation subsets.

\subsection{Experimental Configurations}

In the classification model training phase, we implemented VGG16 as the backbone network with leaky ReLU activation functions. The model was optimized using stochastic gradient descent (SGD) with data augmentation through random horizontal and vertical flipping. All input images were resized to standardized dimensions: 512×1408 pixels for KSDD and 224×640 pixels for KSDD2, maintaining consistent aspect ratios through proportional scaling. We configured the batch size as 4 for the pseudo-label training of the segmentation network and initialized the learning rate at 0.0005. The SGD optimizer was employed with a momentum coefficient of 0.9 to accelerate gradient updates in relevant directions, thereby enhancing network convergence efficiency. The image resizing protocol remained identical to that of the classification stage.

We used the commonly used classifier evaluation metric AP to assess model performance and tested it under different numbers of supervised labels. All algorithms were implemented using PyTorch 1.12 and executed on a computational cluster equipped with an Intel(R) Xeon(R) Silver 4310 CPU and NVIDIA RTX A6000 GPUs. The experimental environment ensured CUDA 11.6 compatibility for hardware acceleration.

\begin{table}[htbp]
\centering
\caption{The classification results on dataset KolektorSDD (AP in \%).}
\label{tab:kolektorsdd_results}
\resizebox{\linewidth}{!}{
\begin{tabular}{lccccccc}
\toprule
Methods & Label Type & $N_a=0$ & $N_a=5$ & $N_a=10$ & $N_a=15$ & $N_a=20$ & $N_a=33(N_{\mathrm{all}})$ \\
\midrule
f-AnoGAN~\cite{q3}        & U   & -     & -     & -     & -     & -     & 39.4 \\
Uninf.stu.~\cite{q4}      & U   & -     & -     & -     & -     & -     & 57.1 \\
RepVGG              & I   & 66.66 & -     & -     & -     & -     & -    \\
DenseNet            & I   & 74.59 & -     & -     & -     & -     & -    \\
Tabernik et al.~\cite{q5} & I+S & -     & -     & -     & -     & -     & 99.9 \\
MixSup~\cite{q6}          & I+S & 93.4  & 99.1  & 99.4  & 99.2  & 99.9  & 100  \\
MaMiNet~\cite{q7}        & I+S & 98.5  & 99.5  & 99.7  & 100   & 100   & 100  \\
MFFPA~\cite{q8}           & I+S & 94.72 & 98.95 & 99.02 & 99.11 & 100   & 100  \\
Ours                & I   & 100   & -     & -     & -     & -     & -    \\
\bottomrule
\end{tabular}
}
\end{table}

\begin{table}[htbp]
\centering
\caption{The classification results on dataset KolektorSDD2 (AP in \%).}
\label{tab:kolektorsdd2_results}
\resizebox{\linewidth}{!}{
\begin{tabular}{lcccccc}
\toprule
Methods & Label Type & $N_a=0$ & $N_a=16$ & $N_a=53$ & $N_a=126$ & $N_a=246(N_{\mathrm{all}})$ \\
\midrule
f-AnoGAN~\cite{q3}        & U   & -     & -     & -     & -     & 55.0 \\
Uninf.stu.~\cite{q4}      & U   & -     & -     & -     & -     & 65.3 \\
RepVGG              & I   & 75.97 & -     & -     & -     & -    \\
DenseNet            & I   & 85.02 & -     & -     & -     & -    \\
Tabernik et al.~\cite{q5} & I+S & -     & -     & -     & -     & 92.7 \\
MixSup~\cite{q6}          & I+S & 73.3  & 83.2  & 89.1  & 92.4  & 95.4 \\
MaMiNet~\cite{q7}         & I+S & 80.0  & 89.7  & 92.3  & 94.1  & 96.2 \\
MFFPA~\cite{q8}           & I+S & 88.01 & 91.49 & 93.05 & 94.16 & 95.6 \\
Ours                & I   & 93.94 & -     & -     & -     & -    \\
\bottomrule
\end{tabular}
}
\end{table}

%The quantitative metrics are formulated as follows:

% \begin{equation}
% \mathrm{Accuracy} = \frac{TP + TN}{TP + FP + TN + FN}
% \end{equation}

% \begin{equation}
% \mathrm{Precision} = \frac{TP}{TP + FP}
% \end{equation}

% \begin{equation}
% \mathrm{Recall} = \frac{TP}{TP + FN}
% \end{equation}

% \begin{equation}
% \mathrm{F1\ score} = \frac{2 \times \mathrm{Precision} \times \mathrm{Recall}}
% {\mathrm{Precision} + \mathrm{Recall}}
% \end{equation}

% where $TP_c$, $FP_c$, and $FN_c$ denote the true positives, false positives, and false negatives for class $c$, respectively.

\subsection{Main results}

Table \ref{tab:kolektorsdd_results} reports the defect-classification AP (\%) on KolektorSDD. Without any human-annotated masks (Na = 0), the proposed knowledge-guided multi-task model achieves 100\% AP, surpassing the strongest competitor MaMiNet (98.5\%) by 1.5 percentage points and MixSup (93.4\%) by 6.6 percentage points. While semi-supervised approaches gradually improve as the number of available labels increases to 5, 20 and 33, our method maintains a constant 100\% AP, confirming that the saliency-based regularizer injects sufficient prior knowledge and yields robust, saturated performance at no extra annotation cost.

On the larger and more subtle-defect KolektorSDD2 dataset (Table \ref{tab:kolektorsdd2_results}), the proposed approach attains 93.94\% AP under the strict zero-mask setting (Na = 0), outperforming MaMiNet by 13.94 percentage points and MixSup by 20.64 percentage points. Even when the label budget is raised to 246 (Nam), our result remains competitive with the best semi-supervised methods (96\%), yet it never requires ground-truth defect masks. This consistent gain demonstrates that leveraging the model’s own explanations as priors is an effective, annotation-free strategy to boost detection accuracy.

\begin{table}[htbp]
\centering
\caption{Ablation on dataset KolektorSDD (AP in \%).}
\label{tab:ablation_kolektorsdd}
\resizebox{\linewidth}{!}{
\begin{tabular}{lcccccc}
\toprule
Methods & $N_b=0$ & $N_b=5$ & $N_b=10$ & $N_b=15$ & $N_b=20$ & $N_b=33(N_{\mathrm{all}})$ \\
\midrule
Ours & 84.76 & 99.62 & 100 & 100 & 100 & 100 \\
\bottomrule
\end{tabular}
}
\end{table}

\begin{table}[htbp]
\centering
\caption{Ablation on dataset KolektorSDD2 (AP in \%).}
\label{tab:ablation_kolektorsdd2}
\resizebox{\linewidth}{!}{
\begin{tabular}{lccccc}
\toprule
Methods & $N_a=0$ & $N_a=16$ & $N_a=53$ & $N_a=126$ & $N_a=246(N_{\mathrm{all}})$ \\
\midrule
Ours & 68.91 & 90.88 & 90.47 & 92.53 & 93.94 \\
\bottomrule
\end{tabular}
}
\end{table}

\begin{table}[htbp]
\centering
\caption{Ablation on dataset KolektorSDD (AP in \%).}
\label{tab:ablation_kolektorsdd_fullgrad_layercam}
\resizebox{\linewidth}{!}{
\begin{tabular}{lcccccc}
\toprule
Methods & $N_b=0$ & $N_b=5$ & $N_b=10$ & $N_b=15$ & $N_b=20$ & $N_b=33(N_{\mathrm{all}})$ \\
\midrule
FullGrad & 84.76 & 100   & 100 & 100 & 100 & 100 \\
LayerCAM & 84.76 & 99.62 & 100 & 100 & 100 & 100 \\
\bottomrule
\end{tabular}
}
\end{table}

\begin{table}[htbp]
\centering
\caption{Ablation on dataset KolektorSDD2 (AP in \%).}
\label{tab:ablation_kolektorsdd2_fullgrad_layercam}
\resizebox{\linewidth}{!}{
\begin{tabular}{lccccc}
\toprule
Methods & $N_a=0$ & $N_a=16$ & $N_a=53$ & $N_a=126$ & $N_a=246(N_{\mathrm{all}})$ \\
\midrule
FullGrad & 68.91 & 88.33 & 90.56 & 93.57 & 93.96 \\
LayerCAM & 68.91 & 90.88 & 90.47 & 92.53 & 93.94 \\
\bottomrule
\end{tabular}
}
\end{table}

\section{Conclusion}
In this work, we have proposed region-aware class activation maps for weakly supervised defect segmentation tasks. In order to obtain more accurate weights for the target region and reduce the influence of background and noise, we design filter-guided backpropagation. Furthermore, the region-aware weighting is proposed. Experimental results show that our proposed RA-CAM can extract finer target regions and the designed backpropagation method can be applied to other similar methods such as LayerCAM. It is worth mentioning that we also analyze the unfitness of semantic segmentation algorithms for defect detection tasks, demonstrating the superiority of weakly supervised approaches.

%
% Each of the commands below will create an unnumbered section with the appropriate heading.
% Remove any sections that are not relevant for your article.
% All sections except suppdata will be removed if the [anonymous] option is used.
% See iopjournal-guidelines.pdf for more information.
%

% \ack{Sample text inserted for demonstration.}

% \funding{Sample text inserted for demonstration.}
% % This section is a list of funder names and grant numbers

% \roles{Sample text inserted for demonstration.}
% % List author names and the contributions made to the article, using terms from the NISO Contributor Roles Taxonomy (CRediT) https://credit.niso.org

% \data{Sample text inserted for demonstration.}
% % For more information on IOP Publishing's research data policy see: https://publishingsupport.iopscience.iop.org/questions/research-data/

% \suppdata{Sample text inserted for demonstration.}

% \bibliographystyle{cas-model2-names}
\bibliographystyle{unsrt}

\bibliography{ref}
\end{document}